%% file: top.tex
\documentclass[runningheads]{llncs}
\usepackage{graphicx}
\usepackage{gensymb}
\usepackage{amsmath,amsfonts}
\usepackage{algorithm2e}
\usepackage[usenames, dvipsnames]{color}
\usepackage{calc}
\usepackage{booktabs}
\usepackage{tabularx}
\usepackage{multirow}

\usepackage{wrapfig}
\usepackage{ifsym}

\usepackage[implicit=false]{hyperref}

\graphicspath{{fig/}}

\input{tex/defs}

\begin{document}
	\title{Enforcing connectivity of 3D linear structures using their 2D projections}
	
	%
	%
	\author{Submission ID: 2651}
	
	\author{Doruk Oner\inst{1\star}\orcidID{0000-0002-9403-4628} \and
		Hussein Osman\inst{1\star}\orcidID{0000-0003-1551-6977} \and \\
		Mateusz Kozi\'nski\inst{2}\orcidID{0000-0002-3187-518X} \and 
		Pascal Fua\inst{1}\orcidID{0000-0002-6702-9970} }
	\authorrunning{D. Oner et al.}
	%
	\institute{
		Computer Vision Laboratory, \'Ecole Polytechnique F\'ed\'erale de Lausanne, Switzerland \\
		\email{doruk.oner@epfl.ch}
		\and Institute of Computer Graphics and Vision, 
		Graz University of Technology, Austria}
	
	%
	%
	%
	%
	\maketitle              
	\renewcommand{\thefootnote}{\fnsymbol{footnote}}
	\footnotetext[1]{The two authors contributed equally to this paper.}
	
	%
	%
	\input{tex/0_abstract.tex}
\input{tex/1_intro.tex}
\input{tex/2_related.tex}

	\input{tex/3_method.tex}
\input{tex/4_experiments.tex}

	\input{tex/5_conclussion.tex}

\input{tex/acknowledgement.tex}
	\bibliographystyle{plain}
	\bibliography{string,vision,learning,biomed,topo,misc}

\end{document}

%% file: tex/defs.tex
\newif\ifdraft
\draftfalse

\definecolor{orange}{rgb}{1,0.5,0}
\definecolor{violet}{RGB}{70,0,170}
\definecolor{magenta}{RGB}{170,0,170}
\definecolor{dgreen}{RGB}{0,150,0}

\usepackage[normalem]{ulem}

\ifdraft
\newcommand{\PF}[1]{{\color{red}{\bf PF: #1}}}

\newcommand{\MK}[1]{{\color{blue}{\bf MK: #1}}}

\newcommand{\DO}[1]{{\color{dgreen}{\bf DO: #1}}}

\newcommand{\HO}[1]{{\color{magenta}{\bf HO: #1}}}

\else
\newcommand{\PF}[1]{}

\newcommand{\MK}[1]{}

\newcommand{\DO}[1]{}
\newcommand{\HO}[1]{}

\fi

\newcommand{\comment}[1]{}

\newcommand{\etal}{~\emph{et al.}}

\DeclareMathOperator{\argmax}{\arg\,\max}

\newcommand{\Topo}{\textnormal{\textsc{TOPO}}}

\newcommand{\p}{\mathbf{p}}
\newcommand{\x}{\mathbf{x}}
\newcommand{\y}{\mathbf{y}}

\newcommand{\gt}{\hat{\y}}

\newcommand{\dst}{{}}

\newcommand{\yd}{\y_\dst}

\newcommand{\ad}{\gt}

\newcommand{\Lc}{L_\mathrm{conn}}

\newcommand{\Ld}{L_\mathrm{disc}}

\newcommand{\MSE}{\textnormal{\textsc{MSE}}}

\newcommand{\omse}{{\bf OURS-3D}}
\newcommand{\oproj}{{\bf OURS-2D}}
\newcommand{\bmse}{{\bf MSE-3D}}
\newcommand{\bproj}{{\bf MSE-2D}}

\newcommand{\UNet}{{\it U-Net}}

\newcommand{\bce}{{\bf CE}}
\newcommand{\bagata}[0]{{\bf Perc}}
\newcommand{\bhomo}{{\bf PHomo}}

\newcommand{\neurons}[0]{{Brain}}
\newcommand{\neuronsmisal}[0]{{Neurons}}
\newcommand{\mra}[0]{{MRA}}

\newcommand{\bneurons}[0]{{\bf Brain}}
\newcommand{\bneuronsmisal}[0]{{\bf Neurons}}
\newcommand{\bmra}[0]{{\bf MRA}}

\newcommand{\APLS}{{\it APLS}}
\newcommand{\TLTS}{{\it TLTS}}

\newcommand{\bAPLS}{{\bf APLS}}
\newcommand{\bTLTS}{{\bf TLTS}}
\newcommand{\bCCQ}{{\bf CCQ}}

%% file: tex/0_abstract.tex

\begin{abstract}

Many biological and medical tasks require the delineation of 3D curvilinear structures such as blood vessels and neurites from image volumes. This is typically done using neural networks trained by minimizing voxel-wise loss functions that do not capture the topological properties of these structures. As a result, the connectivity of the recovered structures is often wrong, which lessens their usefulness. In this paper, we propose to improve the 3D connectivity of our results by minimizing a sum of topology-aware losses on their 2D projections. This suffices to increase the accuracy and to reduce the annotation effort required to provide the required annotated training data. Code is available at \href{https://github.com/doruk-oner/ConnectivityOnProjections}{https://github.com/doruk-oner/ConnectivityOnProjections}.

\keywords{Delineation  \and Neurons \and Microscopy scans \and Topology.}
\end{abstract}

%% file: tex/1_intro.tex
\section{Introduction}

Delineating 3D curvilinear structures, such as veins and arteries visible in computed tomography (CT) scans, or dendrites and axons revealed by light microscopy (LM) scans, is central to many applications. State-of-the-art algorithms typically rely on deep networks trained to classify each voxel as either foreground or background by minimizing a voxel-wise loss. Networks trained this way are good at voxel classification but nevertheless prone to topological errors, such as unwarranted gaps in the linear structures and false interconnections between them. This mostly occurs when vessels and neuronal projections appear as thin but densely woven structures and misclassifying a few voxels can disrupt their connectivity without much influence on voxel-wise accuracy. These errors greatly reduce the usefulness of the resulting arborization models. Correcting them requires manual interventions, which is very time consuming when performed on whole-brain microscopy scans or whole-organ CT scans, especially at scales sufficiently large to produce statistically significant results. 



In other words, networks trained by minimizing losses such as the Cross Entropy and the Mean Squared Error, which are sums of per-voxel terms \emph{independent} of all other voxels, struggle to learn patterns formed \emph{jointly} by groups of voxels~\cite{Hu21b,Clough20,Oner21a}. A promising approach to addressing this issue is to develop {\it topology-aware} loss functions that evaluate patterns emerging from predictions for multiple voxels. This includes perceptual losses~\cite{Mosinska18}, loss functions based on persistent homology~\cite{Hu21b,Clough20}, and a loss that enforces continuity of linear structures by penalizing interconnections between background regions on their opposite sides~\cite{Oner21a}. The latter has proved to be more effective for 2D images than the others but does not naturally generalize to 3D volumes.

In this paper, we start from the observation that continuity of a 3D linear structure implies continuity of its 2D projections. Hence, we can use a topology-aware loss, such as the one of~\cite{Oner21a}, to penalize connectivity errors in 2D projections of the 3D predictions, thereby indirectly penalizing the errors in the 3D originals. This also means that we can use 2D annotations, which are much easier to obtain than full 3D ones, to train a 3D network. This is close in spirit to the approach of~\cite{Kozinski20} in which delineation networks are trained by minimizing a loss function in maximum intensity projections of the predictions and the annotations. 

We demonstrate the effectiveness of our approach for delineating neurons in light microscopy scans and tracing blood vessels in Magnetic Resonance Angiography scans. Not only do we produce topologically correct delineations, but we also reduce the annotation effort required to train the networks. 

%% file: tex/2_related.tex
\section{Related work}
\label{sec:related}

Over the years, many approaches to delineating 3D linear structures have been proposed. They range from hand-designing filters that are sensitive to tubular structures~\cite{Frangi98,Law08,Turetken13c} to learning such filters~\cite{Wu12a,Breitenreicher13} using support vector machines \cite{Huang09}, gradient boost~\cite{Sironi14}, or decision trees~\cite{Turetken16a}. 

Neural networks have now become the dominant technique~\cite{Mnih10,Ganin14,Maninis16,Peng17,Guo19,Wolterink19}. They are often trained by minimizing pixel-wise loss functions, such as the cross-entropy or the mean square error. As a result, the delineations they produce often feature topological mistakes, such as unwarranted gaps or false connections. This occurs because it often takes very few mislabeled pixels to significantly alter the topology with little impact on the pixel-wise accuracy. 

Specialized solutions to this problem have been proposed in the form of loss functions comparing the topology of the predictions to that of the annotations.
For example, the perceptual loss~\cite{Mosinska20} has been shown to be sensitive to topological differences between the prediction and the ground truth, but cannot be guaranteed to penalize all of them.
Persistent Homology~\cite{Edelsbrunner08} is an elegant approach to describing and comparing topological structures. It has been used to define topology-oriented loss functions~\cite{Hu19b,Clough20,Byrne20}. Unfortunately, computing this loss is computationally intensive and error-prone because it does not account for the location of topological structures.
clDice~\cite{Shit21} is a loss function that employs a soft skeletonization algorithm to compare the topology of the prediction and the annotation, but it is designed for volumetric segmentation, whereas we focus on tracing linear structures given their centerlines.

For delineation of 2D road networks, existing approaches are outperformed by the method of~\cite{Oner21a} that repurposes the MALIS loss initially proposed to help better segment electron microscopy scans~\cite{Turaga09,Funke18} to improve the topology of reconstructed loopy curvilinear networks. Unfortunately, the algorithm of~\cite{Oner21a} can only operate in 2D. In this paper, we show how it can nevertheless be exploited for 3D delineation in volumetric images.

%% file: tex/3_method.tex

\section{Approach}

We train a deep network to regress the distance from each voxel of the input 3D image $\x$ to the center of the nearest linear structure. We denote the predicted 3D distance map by  $\yd$. The annotations are given in the form of a graph with nodes in 3D space. We denote the set of edges of this graph by $\mathcal{E}$. From the annotation graph, we compute the truncated ground truth distance map $\ad$. For a voxel $\p$, $\ad[p]=\min((\min_{\epsilon \in \mathcal{E}} d_{p\epsilon}),d_\mathrm{max})$, where $d_{p\epsilon}$ is the distance from $\p$ to the annotation edge $\epsilon$, and $d_{\mathrm{max}}$ is the truncation distance set to  $15$ pixels. 

A simple way to train our deep net is to minimize a Mean Squared Error loss $L_{\MSE}(\yd,\ad)$ for all training images. As discussed in Section~\ref{sec:related}, minimizing such a voxel-wise loss does not guarantee that connectivity is preserved because mislabeling only a few voxels is enough to disrupt it. 
For 2D images, this problem has been addressed with a loss term $L_\Topo$ that effectively enforces continuity of 2D linear structures~\cite{Oner21a}. Unfortunately, it is limited to 2D data by design and cannot be extended to 3D.
To bypass this limitation, we leverage the observation that continuity of 3D structures implies continuity of their 2D projections and evaluate $L_\Topo$ on 2D projections of the 3D predicted and ground truth distance maps.
We introduce the 2D connectivity-oriented loss term and the technique to train 3D deep networks on 2D projections in the following subsections.

\subsection{Connectivity loss}
\label{sec:losscon}
\input{fig/fig_topo}

 In this section, we recall the intuition behind the connectivity-oriented loss term $L_\Topo$ of \cite{Oner21a}. We refer the reader to the original publication for a more detailed explanation. 
As illustrated by Fig.~\ref{fig_topoloss}(a), a path connecting pixels on opposite sides of a linear structure must cross that structure and should therefore contain at least one pixel $\p$ such that the predicted distance map $\yd[\p]=0$. If $\yd$ contains erroneous disconnections, then it is possible to construct a path that connects pixels on opposite sides of the structure, but only crosses pixels with predicted distance values larger than zero, as depicted by Fig.~\ref{fig_topoloss}(b). In particular, a \emph{maximin} path, that is, the path with largest smallest pixel among all possible paths between the same end points, is guaranteed to pass through an interruption of the linear structure, if it exists. $L_\Topo$ minimizes the smallest pixel on the maximin path between each pair of end points that belong to background regions on the opposite sides of annotated linear structures, shown in Fig.~\ref{fig_topoloss}(c). 
It has proven effective in enforcing connectivity of 2D linear structures, but cannot be extended to 3D, because 3D linear structures do not subdivide 3D volumes into disjoint background regions.


\subsection{Projected Connectively Loss}
\label{sec:proj}

\input{tex/fig_projections}

The key observation underlying our approach is that 3D continuity of three-dimensional curvilinear structure, represented as a depth-map, implies its continuity in 2D minimum-intensity projections of the depth map.
The reverse is not true: a projection of a discontinuous 3D depth-map might appear continuous if it is taken along the direction tangent to the linear structure at discontinuity.
However, even in such case, the discontinuity appears in other projections, taken along directions orthogonal to the direction of the first projection, as shown in Fig.~\ref{fig:connectivity_projections}.
In general, given three orthogonal projections of a 3D volume, each discontinuity appears in at least two of them, unless it is occluded by other linear structures. 
Hence, we evaluate the topology-enforcing loss $L_\Topo$ on projections of the predicted and ground truth distance maps along the principal directions.
Let $\yd^i$ be the min-intensity projection of $\yd$ along direction $i$, where $i$ can be one of the axes $x$, $y$, or $z$ and the corresponding projection of $\ad$ be $\ad^i$. We take our connectivity-enforcing loss to be
\begin{equation}
\Lc(\yd,\ad)=\sum_{i\in\{x,y,z\}} L_\Topo(\yd^i,\ad^i) ,
\label{eq:connLoss}
\end{equation} 
where $L_\Topo$ is the 2D connectivity loss of~\cite{Oner21a} discussed above. 
This loss can easily be differentiated with respect to the values of $\yd$, as the minimum-intensity projection is just a column-wise \emph{min} operation.

\subsection{Total Loss}
\label{sec:loss}

The total loss that we minimize can therefore be written as 
\begin{equation}
L_{3D}(\yd, \ad) = L_{\MSE}(\yd,\ad) + \alpha \Lc(\yd,\ad),
\label{eq:fullLoss3D}
\end{equation}
where $\alpha$ is a scalar that weighs the influence of the two terms. As discussed above, $L_{\MSE}$ can be simply computed as the mean squared difference between the predicted and ground truth 3D distance maps. 

This is a perfectly valid choice when 3D annotations are available, but such annotations are typically hard to obtain. Fortunately, it has been shown in~\cite{Kozinski20} that one can train a network to perform 3D volumetric delineation given {\em only} 2D annotations in Maximum Intensity Projections. This saves time because manually delineating in 2D is much easier than in 3D. Since we impose our connectivity constraints on projections along the axes $x$, $y$, and $z$, it makes sense to also provide annotations only for the corresponding projections of the input volume $\x$, and generate from them ground truth distance maps $\ad_x$, $\ad_y$, and $\ad_z$. To replace the 3D ground truth $\ad$, that $L_{3D}$ requires, we can rewrite our total loss as
\begin{align}
L_{2D}(\yd,\ad_x,\ad_y,\ad_z)  & = \sum_{i\in\{x,y,z\}}  L_\MSE (\yd^i,\ad_i)+ \alpha \sum_{i\in\{x,y,z\}} L_\Topo (\yd^i,\ad_i), 
\label{eq:fullLoss2D}
\end{align}
where the Mean Squared Error is evaluated on the minimum-intensity projections of the predicted distance map and the distance map produced for the 2D annotation of data projection.

%% file: fig/fig_topo.tex

\begin{figure*}[hbt!]
\centering
\vspace{-3mm}
\setlength{\tabcolsep}{6pt}
\begin{tabular}{@{} >{\centering\arraybackslash}m{0.30\textwidth} >{\centering\arraybackslash}m{0.30\textwidth} >{\centering\arraybackslash}m{0.30\textwidth} @{}}
	\includegraphics[width=0.20\textwidth]{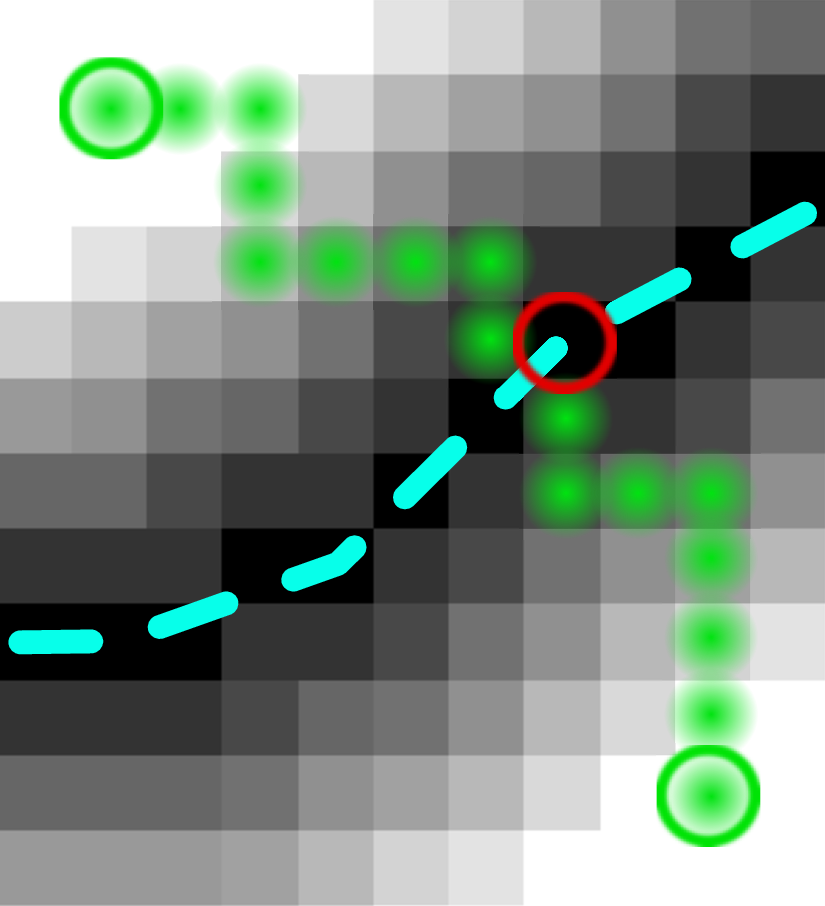}&
	\includegraphics[width=0.20\textwidth]{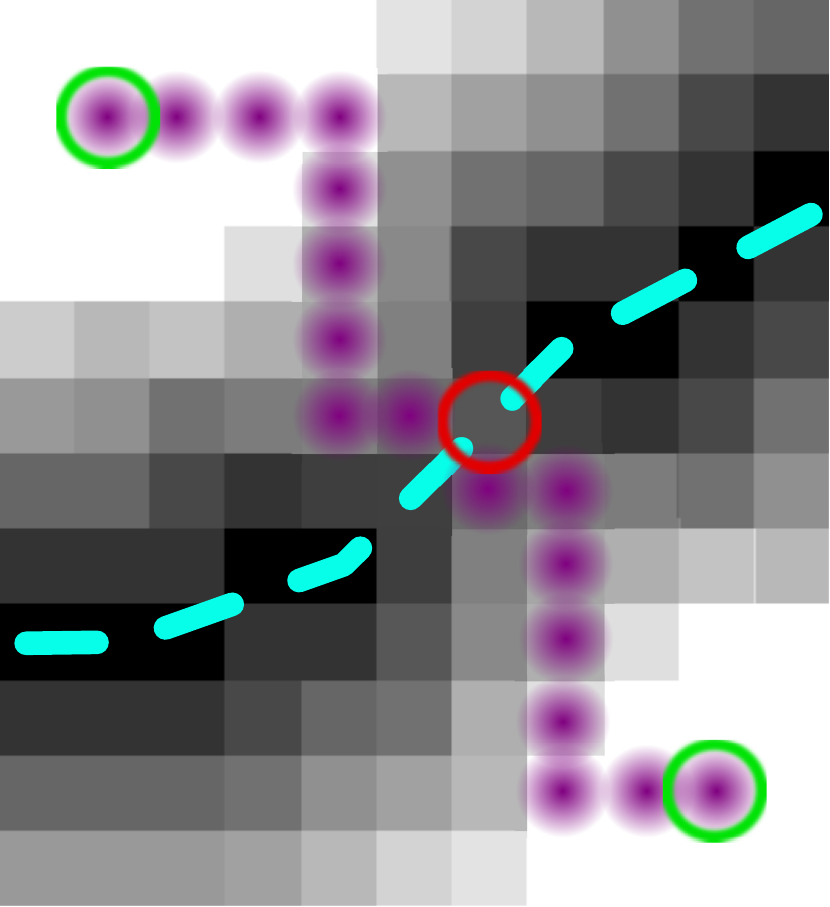} &
	\includegraphics[width=0.20\textwidth]{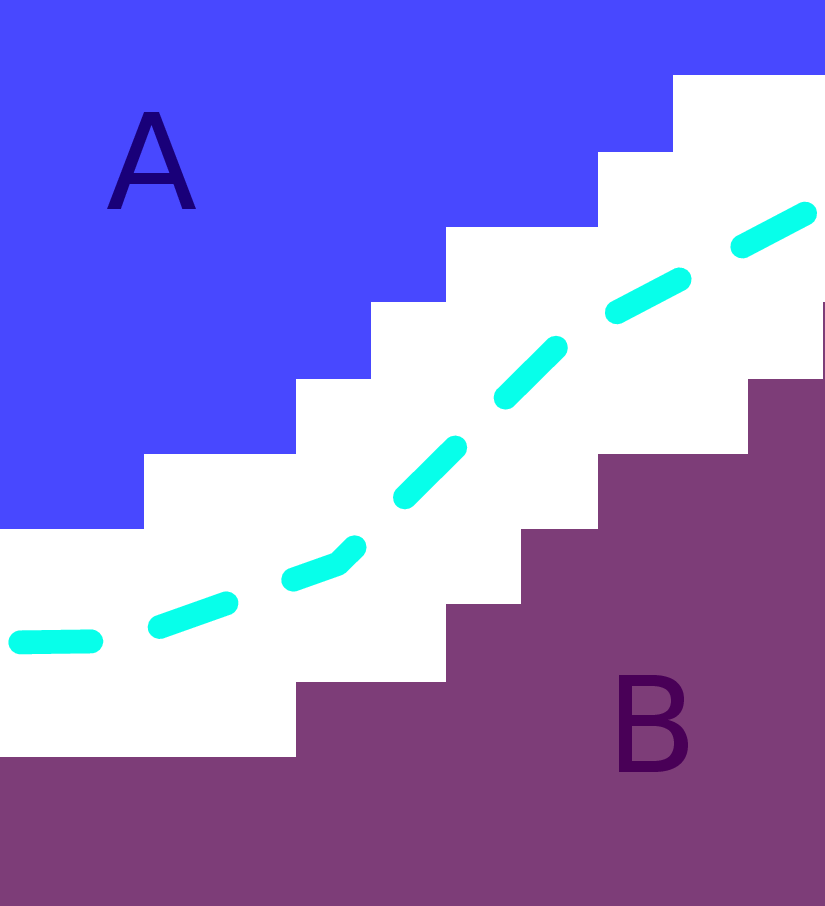} \\
	(a) & (b) & (c)
\end{tabular}
\caption{
The intuition behind $L_\Topo$. {\it (a)} In a perfect distance map, any path connecting pixels on the opposite sides of an annotation line (dashed, magenta) crosses a zero-valued pixel (red circle). {\it (b)} If a distance map has erroneously high-valued pixels along the annotation line, the maximin path (violet) between the same pixels crosses one of them (red circle). 
\emph{(c)} The connectivity-oriented loss $L_\Topo$ is a sum of the smallest values crossed by maximin paths connecting pixels from different background regions. The background regions are computed by first dilating the annotation (dilated annotation shown in white), to accommodate possible annotation inaccuracy.
\label{fig_topoloss}
}
\end{figure*}

%% file: tex/fig_projections.tex
\begin{wrapfigure}{R}{0.3\textwidth}
\centering
\includegraphics[width=0.3\textwidth]{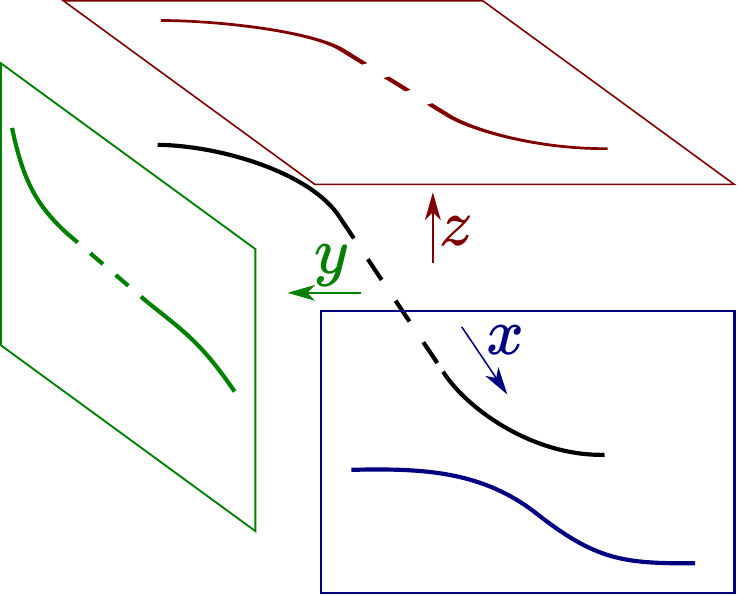}
\caption{
\label{fig:connectivity_projections}
Disconnections in 3D linear structures appear in at least two out of three orthogonal projections, unless the structure is occluded.
}
\end{wrapfigure}

%% file: tex/4_experiments.tex
\section{Experiments}

\subsection{Datasets}
\label{sec:datasets}

We tested our approach on three data sets. The \bneurons{} 
comprises 14 light microscopy scans of mouse brain, sized $250\times 250\times 250$. We use 10 of them for training and 4 as a validation test.
\bneuronsmisal{} contains 13 light microscopy scans of mouse neurons, sized $216\times 238\times 151$.
We use 10 for training and 3 for validation. 
\bmra{} is a publicly available set of Magnetic Resonance Angiography brain scans~\cite{Bullitt05}.
We crop them to size $416\times 320\times 28$ by removing their empty margins, and use 31 annotated scans for training and 11 for validation.
A sample image from each data set can be found in Fig.~\ref{fig_results}. 

\subsection{Metrics}
\label{sec:metrics}

We use the following performance metrics. \bCCQ{}~\cite{Wiedemann98}, {\it correctness}, {\it completeness}, and {\it quality} are similar to precision, recall, and the F1 score, but predicted foreground voxels are counted as true positives if they are closer than 3 voxels away from the ground truth ones. \bAPLS{}~\cite{VanEtten18} is defined as the mean of relative length differences between shortest paths connecting corresponding pairs of randomly selected end points in the ground truth and predicted graphs. \bTLTS{} ~\cite{Wegner13} is the fraction of shortest paths in the prediction that are less than 15\% longer or shorter than the corresponding ground truth paths.

\input{tab/results}

\input{fig/fig_boxplot_all}

\input{fig/fig_results}

\subsection{Architectures and Baselines.}

In all experiments, we use a 3D U-Net~\cite{Ronneberger15} with three max-pooling layers and two convolutional blocks. The first layer has 32 filters. Each convolution is followed by a batch-norm and dropout with a probability of 0.1. We used a batch size of 4. For data augmentation, we randomly crop volumes of size $96\times96\times96$ and flip them over the three axes. The networks were trained for $50k$ iterations with Adam~\cite{Kingma15}, with the learning rate of $1e-3$ and weight decay of $1e-3$. At test time, the predicted distance map is thresholded at 2 and skeletonized to obtain centerlines. To compute the \TLTS{} and \APLS{}, we extract graphs from the prediction and the ground-truth, based on voxel connectivity. 

As discussed in Section~\ref{sec:loss}, we can train our network by minimizing either $L_{3D}$~\eqref{eq:fullLoss3D} or $L_{2D}$~\eqref{eq:fullLoss2D}. Recall that computing  $L_{3D}$ requires 3D annotations, while 2D annotations suffice to compute $L_{2D}$. We will refer to these approaches as \omse{} and \oproj{}. We compare the results we obtain in this way to:
%
\begin{itemize}
	\item \bmse{}. $L_{MSE}$ between 3D predictions and ground truths.
	
	\item \bproj{}. $L_{MSE}$ between 2D ground truth and projected predictions~\cite{Kozinski20}. 
	
	\item \bce{}. 3D binary segmentation trained with Cross-Entropy (CE).
	
	\item \bagata{}. A weighted sum of CE and a perceptual loss function that compares feature maps computed for the ground truth and predicted distance maps~\cite{Mosinska18}. To extract the feature maps, we use a ResNet50 architecture pre-trained with 23 different biomedical datasets~\cite{Chen19d}. 
	
	\item \bhomo{}. A weighted sum of CE and a loss based on Persistent Homology~\cite{Hu19b}. 
	
\end{itemize}
For $\bagata$ and $\bhomo{}$, we use the weighing coefficients recommended in the original publications. For \omse{} and \oproj{}, $\alpha$ is set to $1e-3$ and $\beta$ to $0.1$. These values are selected empirically, based on the ablation study we provide in the supplementary material. We used windows of size 48 pixel to calculate $L_\Topo$.

\subsection{Results.} 
\label{subs:results}

We provide qualitative results in Fig.~\ref{fig_results} and quantitative ones in Table~\ref{tab:results}. Minimizing our loss function consistently improves the \bAPLS{} and \bTLTS{} by a significant margin compared to minimizing pixel-wise losses.  Additionally, our loss outperforms the topology-aware losses \bagata{} and \bhomo{}. It also delivers a boost, albeit only on average, in terms of the \bCCQ{}. Box plots in Fig.~\ref{fig_boxplot}, show that these conclusions hold when variance of the scores is taken into account. 

On average, \oproj{} achieves performance that are slightly lower than those of comparable performance to \omse{}. However, annotating 2D slices instead of 3D stacks significantly reduces the time required to annotate, as shown in the user study conducted in~\cite{Kozinski18}. Thus, when annotation effort is a concern, \oproj{} is an excellent alternative to \omse{}. 



%% file: tab/results.tex

\begin{table}[t]
\centering
\caption{
Comparative results.  
A \UNet{} trained with our loss function outperforms existing methods by a considerable margin in terms of the topology-aware metrics.  The improvement in terms of the pixel-wise metrics is smaller but still there on average.
\label{tab:results}
}
\begin{small}
\setlength{\tabcolsep}{5pt}
\begin{tabular}{@{} l l c c   c   c  c  c@{} }
\toprule
 && \multicolumn{3}{c}{Pixel-wise} & &  \multicolumn{2}{c}{Topology-aware}\\
 \cmidrule{3-5}
 \cmidrule{7-8}
Dataset  &  Methods  &     
Corr. &      Comp. &       Qual.  &  &      APLS &        TLTS \\
\midrule
\multirow{7}{*}{\bneurons{}}
						&\bmse{} &
        							96.6 &        93.5 &        90.5 &   &     77.4 &       81.7 \\ 
						&\bproj{} &
        							{\bf 98.3} &        93.6 &        92.1 &  &      76.2 &        80.1 \\
						&\bce{} &
        							97.3 &        96.7 &      94.2     &   &     71.0 &        81.2 \\
        				&\bagata{} &
        							97.6&        96.7 &        94.5 &    &    76.6 &        84.1 \\
						&\bhomo{} &
							        97.5 &        {\bf 96.9} &        94.7 &    &    81.5 &        83.9 \\
						&\omse{} &
        							{\bf 98.3} & 96.7 &{\bf 95.1} & & 87.1 &{\bf 89.6} \\
						&\oproj{}  &
									97.8 & 96.3 & 94.3 & &{\bf 91.6} & 87.4 \\
\midrule
\multirow{7}{*}{\bneuronsmisal{}}
						&\bmse{}&
									80.6 &        83.5 &        69.5 &   &     62.9 &       69.1 \\ 
						&\bproj{} &
									78.4 &        83.5 &        67.9 &  &      65.6 &        71.8 \\
						&\bce{} &
									79.5 &        82.6 &        68.1 &  &      61.2 &        68.6 \\
						&\bagata{}  &
									80.1 &        85.0 &        70.2 &  &      68.9 &        74.5 \\
						&\bhomo{} &
									{\bf 81.3} &        84.8 &        {\bf 71.0} &  &      69.4 &        75.2 \\
						&\omse{} &
									79.9 &        {\bf 86.4} &        {\bf 70.9} &  &      75.1 &        80.2 \\
						&\oproj{} &
									80.3 &        85.5 &        70.7 &  &      {\bf 76.3} &        {\bf 81.2} \\
\midrule
\multirow{7}{*}{\bmra{}}
						&\bmse{} &
									84.9 &        81.2 &        70.8 &  &      58.5 &        60.4 \\
						&\bproj{} &
									83.0 &        82.3 &        70.3 &  &      58.7 &        59.6 \\
						& \bce{} &
									{\bf 85.7} &        81.1 &        71.3 &  &      58.8 &        60.0 \\
						&\bagata{} &
									83.4 &        83.9 &        71.9 &  &      60.9 &        64.5 \\
						&\bhomo{} &
									85.3 &        83.5 &        72.8 &  &      62.1 &        65.2 \\
						&\omse{} &
									81.5 &        {\bf 89.5} &        {\bf 74.3} &  &   {\bf 70.7}    &        {\bf 72.0} \\
						&\oproj{} &
									80.3 &        87.3 &        71.8 &  &      70.5 &        {\bf 71.9} \\
\bottomrule
\end{tabular}
\end{small}
\end{table}

%% file: fig/fig_boxplot_all.tex

\begin{figure*}[hbt!]
\centering
\setlength{\tabcolsep}{3pt}
\begin{tabular}{c c c c}
	\rotatebox[origin=lb]{90}{\parbox{0.22\textwidth}{\centering\neurons{}}} &
	\includegraphics[width=0.3\textwidth]{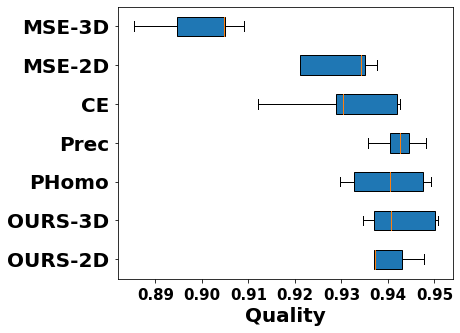}&
	\includegraphics[width=0.3\textwidth]{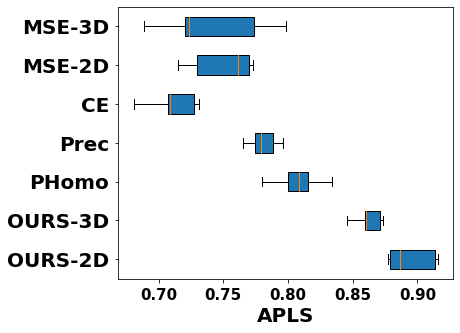} &
	\includegraphics[width=0.3\textwidth]{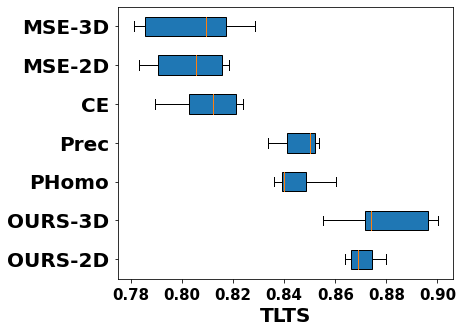} \\
	\rotatebox[origin=lb]{90}{\parbox{0.22\textwidth}{\centering\neuronsmisal{}}} &
	\includegraphics[width=0.3\textwidth]{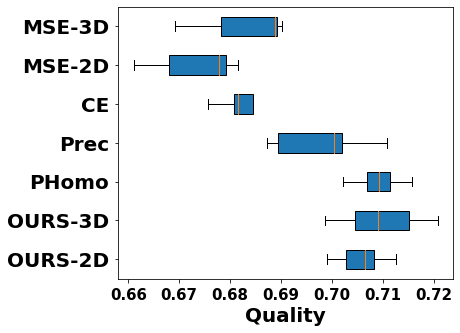}&
	\includegraphics[width=0.3\textwidth]{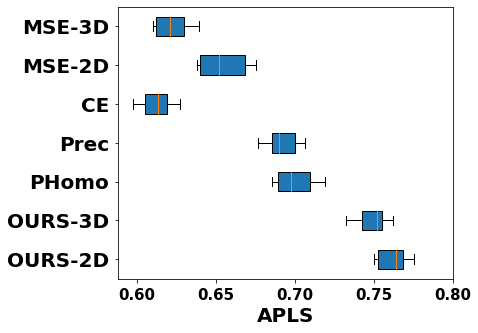} &
	\includegraphics[width=0.3\textwidth]{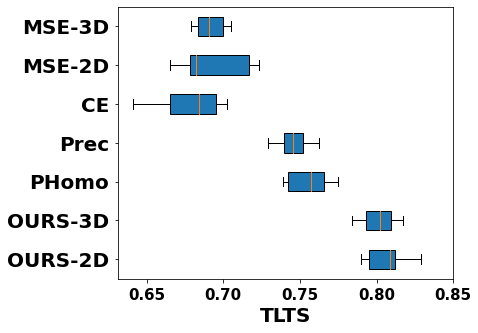} \\
	\rotatebox[origin=lb]{90}{\parbox{0.22\textwidth}{\centering\mra{}}} &
	\includegraphics[width=0.3\textwidth]{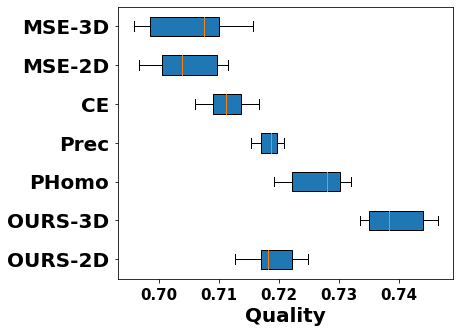}&
	\includegraphics[width=0.3\textwidth]{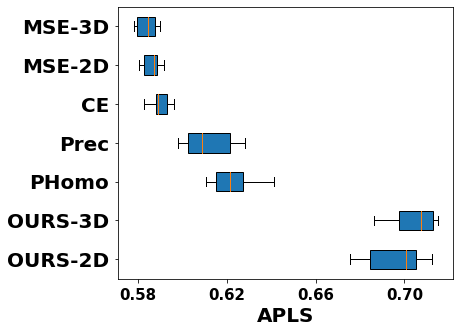} &
	\includegraphics[width=0.3\textwidth]{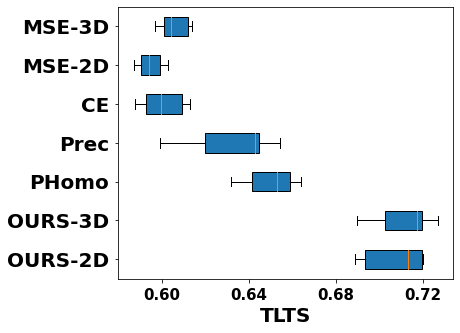} 
\end{tabular}
\caption{Median and quartiles over five training runs of scores attained by networks trained with different loss functions. Minimizing our topology-aware loss results in significantly higher score values than minimizing hte baselines. }
\label{fig_boxplot}
\end{figure*}

%% file: fig/fig_results.tex

\begin{figure*}[hbt!]
\centering
\setlength{\tabcolsep}{3pt}
\begin{tabular}{c c c c c}
	& \bneurons{} & \bneuronsmisal{} & \bneuronsmisal{} & \bmra{} \\
	\raisebox{11mm}{\rotatebox[origin=t]{90}{{\bf Input}}} &
	\includegraphics[width=0.20\textwidth]{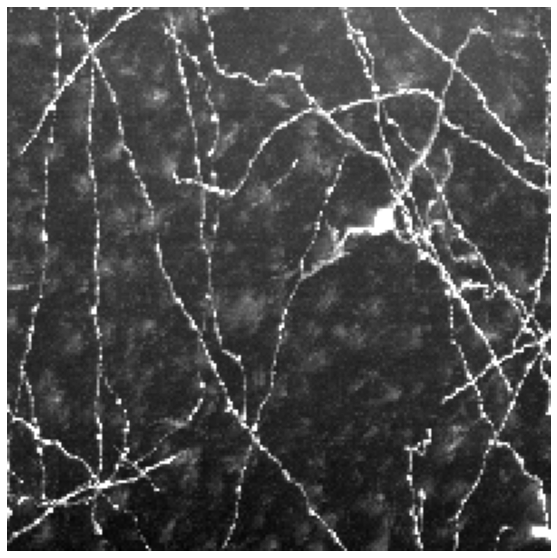}&
	\includegraphics[width=0.20\textwidth]{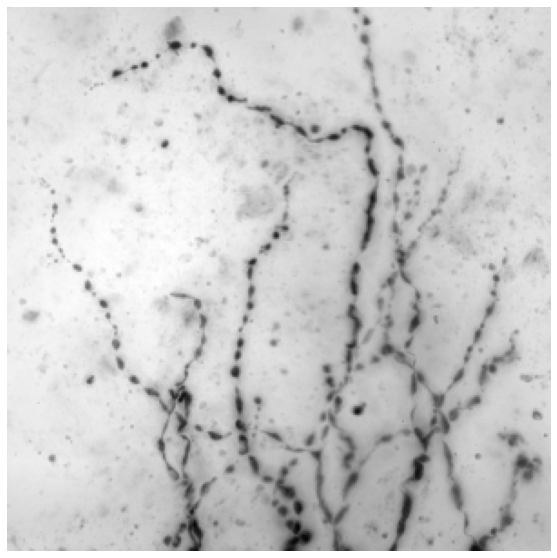} &
	\includegraphics[width=0.20\textwidth]{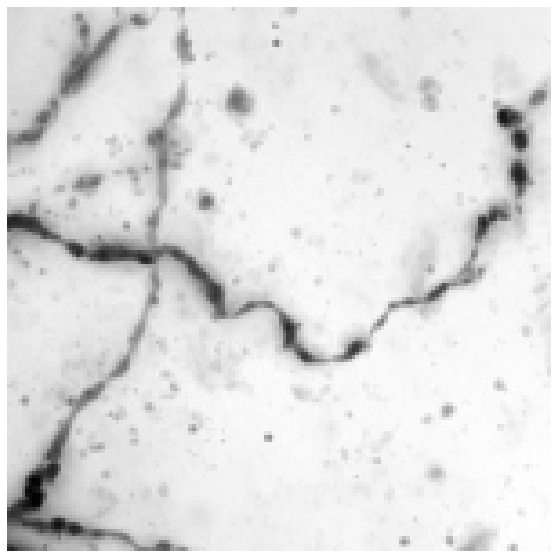} &
	\includegraphics[width=0.20\textwidth]{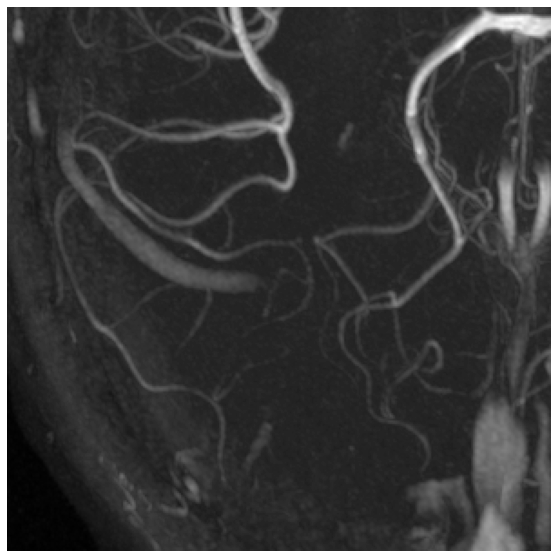}\\
	
	\raisebox{11mm}{\rotatebox[origin=t]{90}{\bagata{}}} &
	\includegraphics[width=0.20\textwidth]{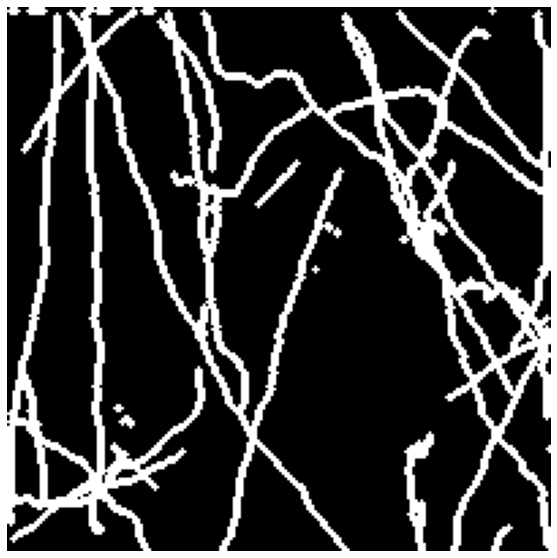}&
	\includegraphics[width=0.20\textwidth]{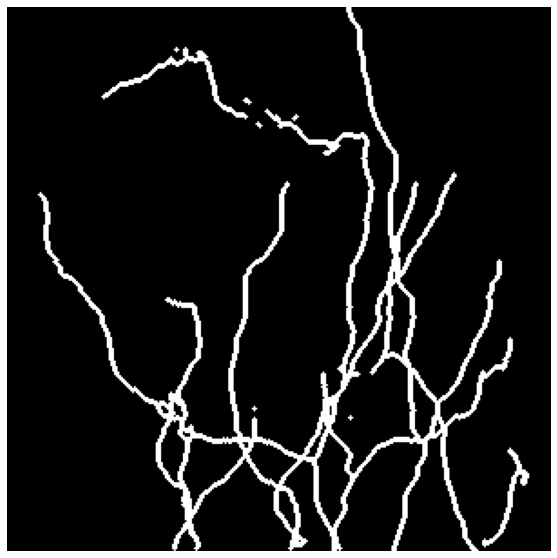} &
	\includegraphics[width=0.20\textwidth]{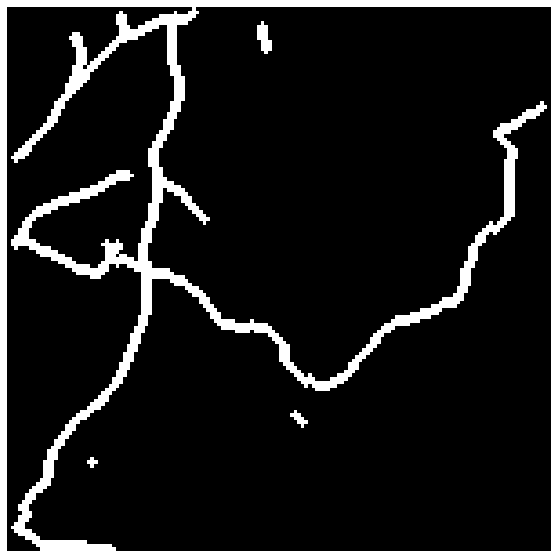} &
	\includegraphics[width=0.20\textwidth]{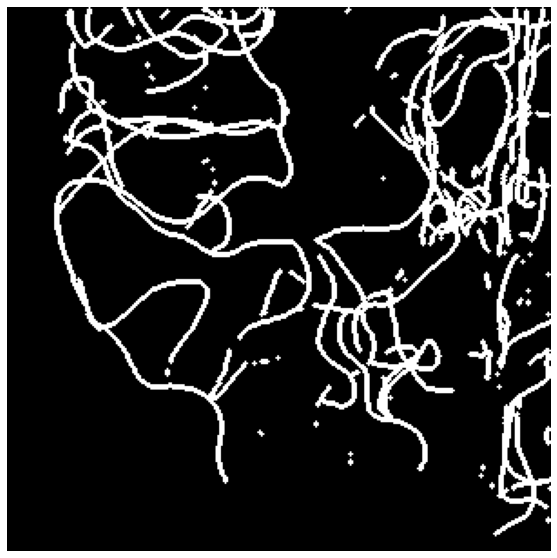} \\	
	
	\raisebox{11mm}{\rotatebox[origin=t]{90}{\bhomo{}}} &
	\includegraphics[width=0.20\textwidth]{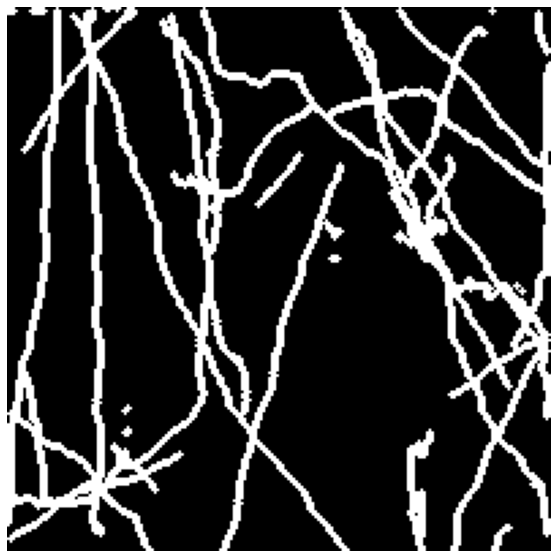}&
	\includegraphics[width=0.20\textwidth]{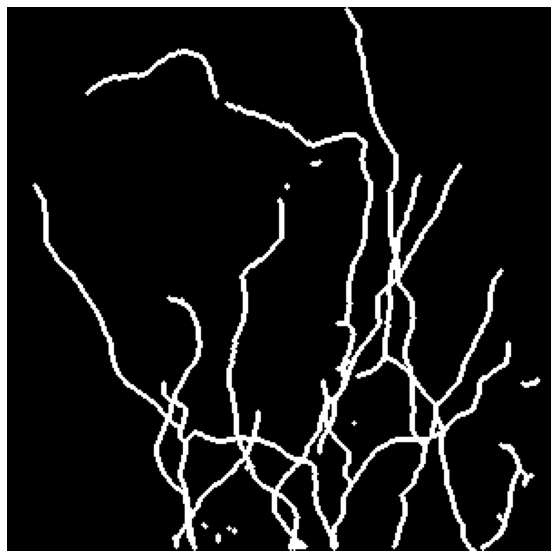} &
	\includegraphics[width=0.20\textwidth]{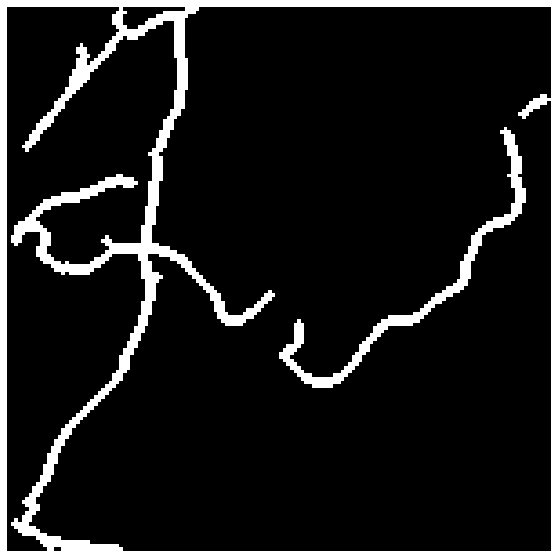} &
	\includegraphics[width=0.20\textwidth]{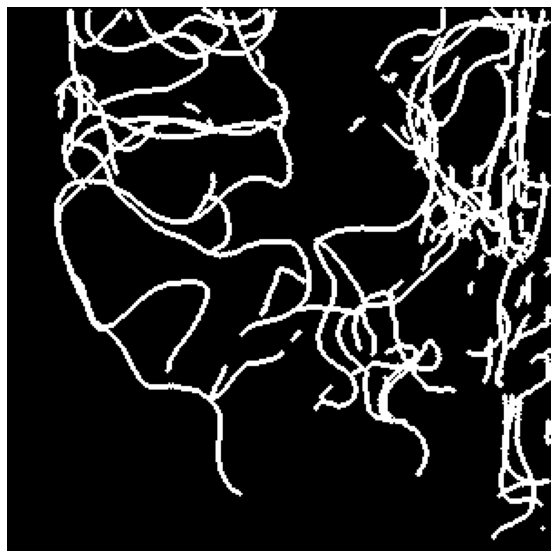} \\	
	
	\raisebox{11mm}{\rotatebox[origin=t]{90}{\omse{}}} &
	\includegraphics[width=0.20\textwidth]{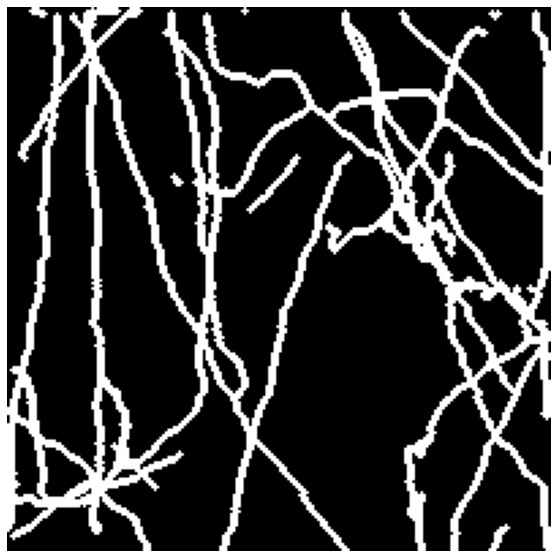}&
	\includegraphics[width=0.20\textwidth]{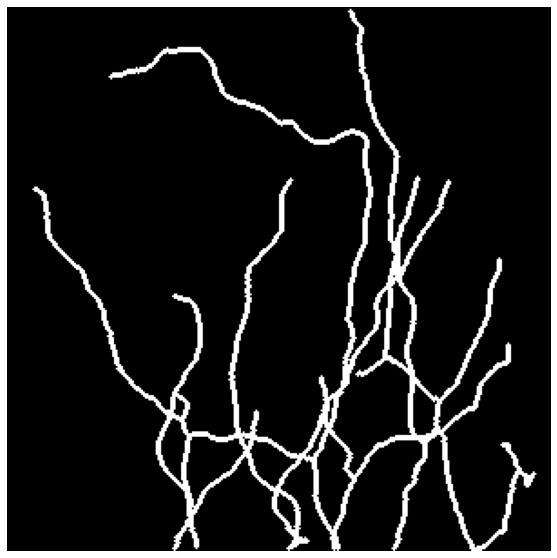} &
	\includegraphics[width=0.20\textwidth]{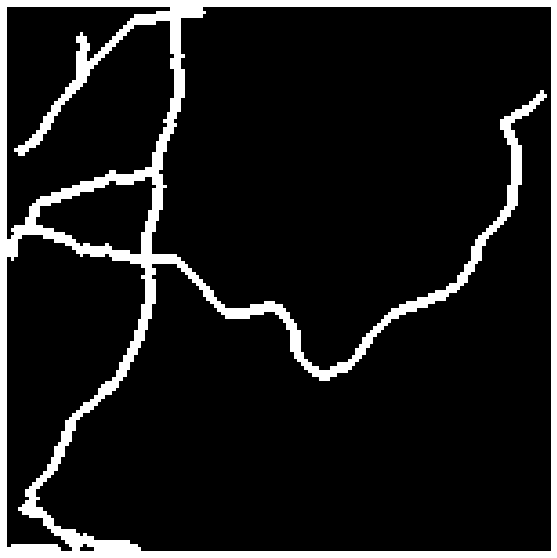} &
	\includegraphics[width=0.20\textwidth]{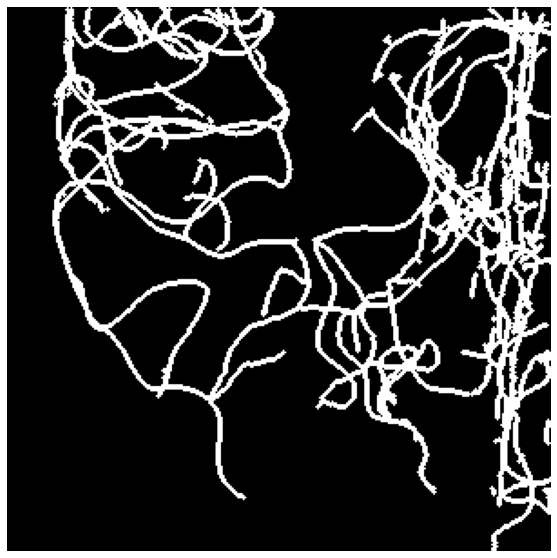} \\	
	
	\raisebox{11mm}{\rotatebox[origin=t]{90}{\oproj{}}} &
	\includegraphics[width=0.20\textwidth]{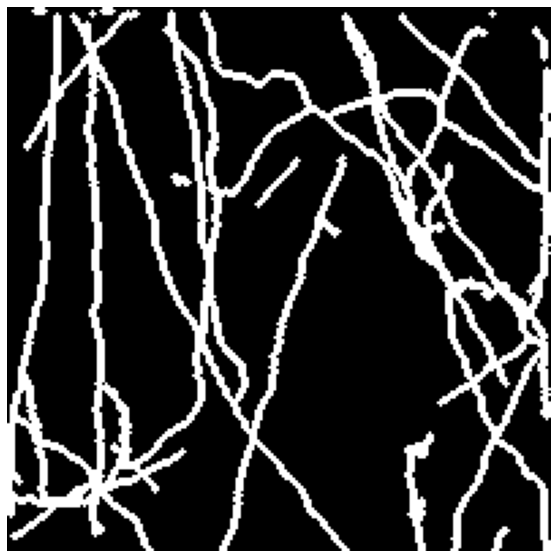}&
	\includegraphics[width=0.20\textwidth]{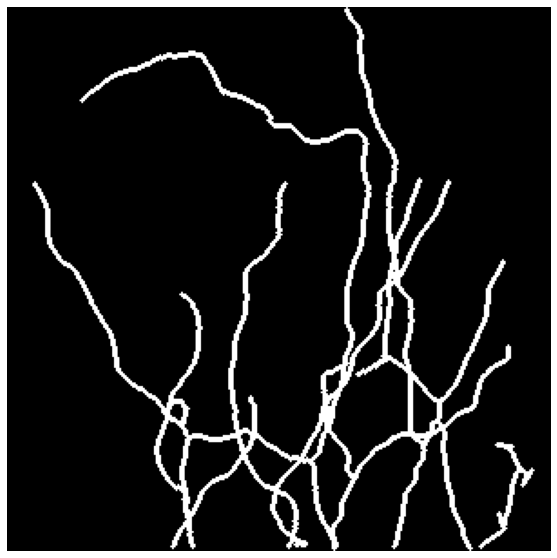} &
	\includegraphics[width=0.20\textwidth]{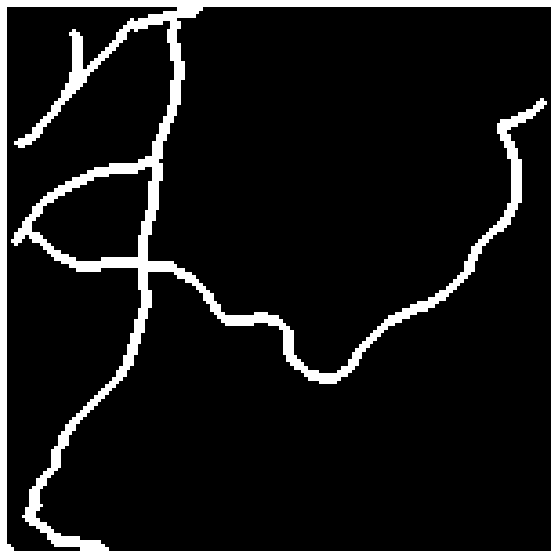} &
	\includegraphics[width=0.20\textwidth]{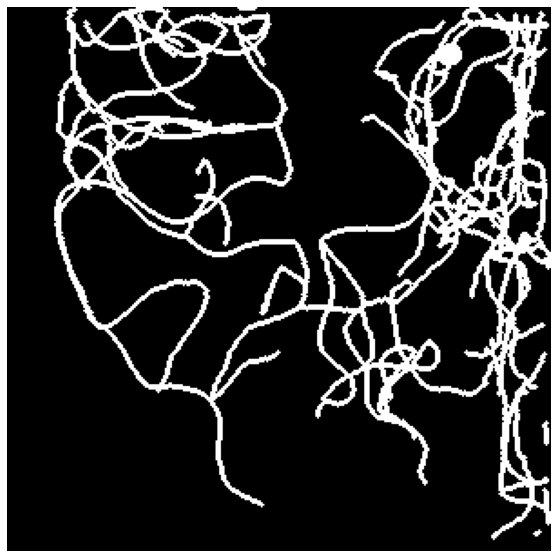} \\	
	
	\raisebox{11mm}{\rotatebox[origin=t]{90}{{\bf Annotation}}} &
	\includegraphics[width=0.20\textwidth]{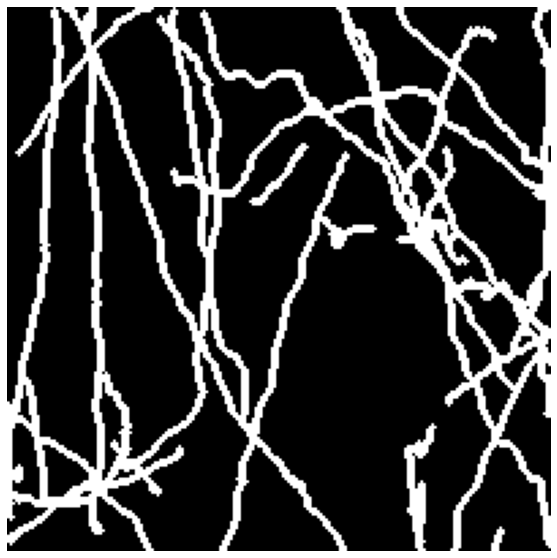}&
	\includegraphics[width=0.20\textwidth]{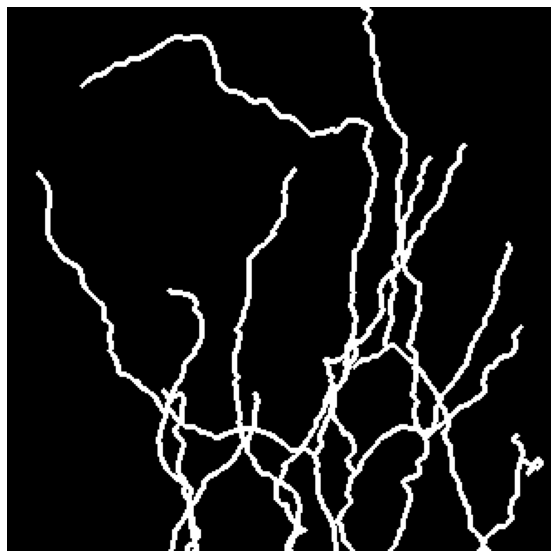} &
	\includegraphics[width=0.20\textwidth]{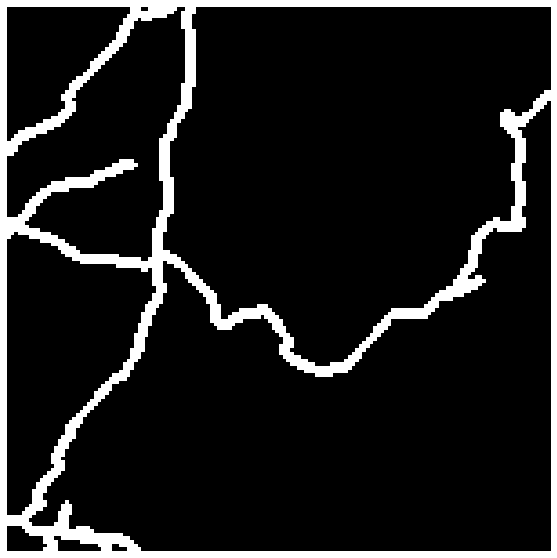} &
	\includegraphics[width=0.20\textwidth]{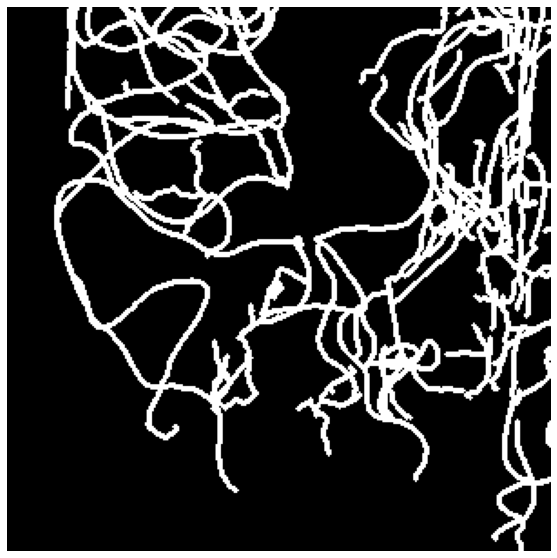}\\
		
\end{tabular}
\caption{
Qualitative comparison of the test results in three different datasets. The connectivity improves significantly when our approach is used.
\label{fig_results}
}
\end{figure*}

%% file: tex/5_conclussion.tex
\section{Conclusion and Future Work}
We proposed a loss function that enforces topological consistency in 2D projections. Training a deep net with our loss greatly improves the 3D connectivity of its outputs and reduces the annotation effort required to obtain training data. In our current implementation, we use projection direction independently of the shape of the delineated structures. However, some projections are more informative than others. To further improve delineation accuracy while reducing the required annotation effort, we will develop algorithms for automatic selection of the optimal projection direction for different parts of the volume, so that we can use less than three projections. 

%% file: tex/acknowledgement.tex
\subsubsection*{Acknowledgements.}

D. Oner received support from the Swiss National Science Foundation under Sinergia grant number 177237. 
M. Kozinski was supported by the FWF Austrian Science Fund Lise Meitner grant no.\ M3374.

%% file: top.bbl
\begin{thebibliography}{10}

\bibitem{Breitenreicher13}
D.~Breitenreicher, M.~Sofka, S.~Britzen, and S.K. Zhou.
\newblock {Hierarchical Discriminative Framework for Detecting Tubular
  Structures in {3D} Images}.
\newblock In {\em Conference on Medical Image Computing and Computer Assisted
  Intervention}, pages 328--340, 2013.

\bibitem{Turaga09}
K.~Briggman, W.~Denk, S.~Seung, M.~Helmstaedter, and S.~Turaga.
\newblock {Maximin Affinity Learning of Image Segmentation}.
\newblock In {\em Advances in Neural Information Processing Systems}, pages
  1865--1873, 2009.

\bibitem{Bullitt05}
E.~Bullitt, D.~Zeng, G.~Gerig, S.~Aylward, S.~Joshi, J.~Smith, W.~Lin, and
  M.~Ewend.
\newblock {Vessel Tortuosity and Brain Tumor Malignancy: A Blinded Study}.
\newblock {\em Acad Radiol}, 12(10):1232--1240, October 2005.

\bibitem{Byrne20}
N.~Byrne, J.~Clough, G.~Montana, and A.~King.
\newblock A persistent homology-based topological loss function for multi-class
  {CNN} segmentation of cardiac {MRI}.
\newblock In {\em STACOM Workshop at MICCAI2020}, volume 12592 of {\em Lecture
  Notes in Computer Science}, pages 3--13. Springer, 2020.

\bibitem{Chen19d}
S.~Chen, K.~Ma, and Y.~Zheng.
\newblock {Med3D: Transfer Learning for 3D Medical Image Analysis}.
\newblock In {\em arXiv Preprint}, 2019.

\bibitem{Clough20}
J.~Clough, N.~Byrne, I.~Oksuz, V.A. Zimmer, J.A. Schnabel, and A.~King.
\newblock A topological loss function for deep-learning based image
  segmentation using persistent homology.
\newblock {\em IEEE Transactions on Pattern Analysis and Machine Intelligence},
  2020.

\bibitem{Edelsbrunner08}
H.~Edelsbrunner and J.~Harer.
\newblock Persistent homology - a survey.
\newblock {\em Contemporary mathematics}, 453:257--282, 2008.

\bibitem{VanEtten18}
A.~Van Etten, D.~Lindenbaum, and T.~Bacastow.
\newblock {Spacenet: {A} Remote Sensing Dataset and Challenge Series}.
\newblock {\em arXiv Preprint}, 2018.

\bibitem{Frangi98}
A.F. Frangi, W.J. Niessen, K.L. Vincken, and M.A. Viergever.
\newblock {Multiscale Vessel Enhancement Filtering}.
\newblock {\em Lecture Notes in Computer Science}, 1496:130--137, 1998.

\bibitem{Funke18}
J.~{Funke}, F.~D. {Tschopp}, W.~{Grisaitis}, A.~{Sheridan}, C.~{Singh},
  S.~{Saalfeld}, and S.~C. {Turaga}.
\newblock {Large Scale Image Segmentation with Structured Loss Based Deep
  Learning for Connectome Reconstruction}.
\newblock {\em IEEE Transactions on Pattern Analysis and Machine Intelligence},
  41(7):1669--1680, 2018.

\bibitem{Ganin14}
Y.~Ganin and V.~Lempitsky.
\newblock {N4-Fields: Neural Network Nearest Neighbor Fields for Image
  Transforms}.
\newblock In {\em Asian Conference on Computer Vision}, pages 536--551, 2014.

\bibitem{Guo19}
Z.~Guo, J.~Bai, Y.~Lu, X.~Wang, K.~Cao, Q.~Song, M.~Sonka, and Y.~Yin.
\newblock {Deepcenterline: {A} Multi-Task Fully Convolutional Network for
  Centerline Extraction}.
\newblock In {\em {IPMI}}, pages 441--453, 2019.

\bibitem{Hu19b}
X.~Hu, F.~Li, D.~Samaras, and C.~Chen.
\newblock {Topology-Preserving Deep Image Segmentation}.
\newblock In {\em Advances in Neural Information Processing Systems}, pages
  5658--5669, 2019.

\bibitem{Hu21b}
X.~Hu, Y.~Wang, L.~Fuxin, D.~Samaras, and C.~Chen.
\newblock {Topology-Aware Segmentation Using Discrete Morse Theory}.
\newblock In {\em International Conference on Learning Representations}, 2021.

\bibitem{Huang09}
X.~Huang and L.~Zhang.
\newblock {Road Centreline Extraction from High-Resolution Imagery Based on
  Multiscale Structural Features and Support Vector Machines}.
\newblock {\em International Journal of Remote Sensing}, 30:1977--1987, 2009.

\bibitem{Kingma15}
D.~P. Kingma and J.~Ba.
\newblock {Adam: {A} Method for Stochastic Optimisation}.
\newblock In {\em International Conference on Learning Representations}, 2015.

\bibitem{Kozinski18}
M.~Kozi{\'n}ski, A.~Mosi{\'n}ska, M.~Salzmann, and P.~Fua.
\newblock {Learning to Segment 3D Linear Structures Using Only 2D Annotations}.
\newblock In {\em Conference on Medical Image Computing and Computer Assisted
  Intervention}, pages 283--291, 2018.

\bibitem{Kozinski20}
M.~Kozi{\'n}ski, A.~Mosinska, M.~Salzmann, and P.~Fua.
\newblock {Tracing in 2D to Reduce the Annotation Effort for 3D Deep
  Delineation of Linear Structures}.
\newblock {\em Medical Image Analysis}, 60, 2020.

\bibitem{Law08}
M.~Law and A.~Chung.
\newblock {Three Dimensional Curvilinear Structure Detection Using Optimally
  Oriented Flux}.
\newblock In {\em European Conference on Computer Vision}, pages 368--382,
  2008.

\bibitem{Maninis16}
K.K. Maninis, J.~Pont-Tuset, P.~Arbel\'{a}ez, and L.~Van Gool.
\newblock {Deep Retinal Image Understanding}.
\newblock In {\em Conference on Medical Image Computing and Computer Assisted
  Intervention}, pages 140--148, 2016.

\bibitem{Mnih10}
V.~Mnih and G.E. Hinton.
\newblock {Learning to Detect Roads in High-Resolution Aerial Images}.
\newblock In {\em European Conference on Computer Vision}, pages 210--223,
  2010.

\bibitem{Mosinska20}
A.~Mosi{\'n}ska, M.~Kozinski, and P.~Fua.
\newblock {Joint Segmentation and Path Classification of Curvilinear
  Structures}.
\newblock {\em IEEE Transactions on Pattern Analysis and Machine Intelligence},
  42(6):1515--1521, 2020.

\bibitem{Mosinska18}
A.~Mosi{\'n}ska, P.~Marquez-Neila, M.~Kozinski, and P.~Fua.
\newblock {Beyond the Pixel-Wise Loss for Topology-Aware Delineation}.
\newblock In {\em Conference on Computer Vision and Pattern Recognition}, pages
  3136--3145, 2018.

\bibitem{Oner21a}
D.~Oner, M.~Kozi{\'n}ski, L.~Citraro, N.~C. Dadap, A.~G. Konings, and P.~Fua.
\newblock {Promoting Connectivity of Network-Like Structures by Enforcing
  Region Separation}.
\newblock {\em IEEE Transactions on Pattern Analysis and Machine Intelligence},
  2021.

\bibitem{Peng17}
H.~Peng, Z.~Zhou, E.Meijering, T.Zhao, G.A. Ascoli, and M.Hawrylycz.
\newblock {Automatic Tracing of Ultra-Volumes of Neuronal Images}.
\newblock {\em Nature Methods}, 14:332--333, 2017.

\bibitem{Ronneberger15}
O.~Ronneberger, P.~Fischer, and T.~Brox.
\newblock {{U-Net}: Convolutional Networks for Biomedical Image Segmentation}.
\newblock In {\em Conference on Medical Image Computing and Computer Assisted
  Intervention}, pages 234--241, 2015.

\bibitem{Shit21}
S.~Shit, J.~Paetzold, A.~Sekuboyina, I.~Ezhov, A.~Unger, A.~Zhylka, J.~Pluim,
  U.~Bauer, and B.~Menze.
\newblock cldice - {A} novel topology-preserving loss function for tubular
  structure segmentation.
\newblock In {\em CVPR}, pages 16560--16569. Computer Vision Foundation /
  {IEEE}, 2021.

\bibitem{Sironi14}
A.~Sironi, V.~Lepetit, and P.~Fua.
\newblock {Multiscale Centerline Detection by Learning a Scale-Space Distance
  Transform}.
\newblock In {\em Conference on Computer Vision and Pattern Recognition}, 2014.

\bibitem{Turetken13c}
E.~Turetken, C.~Becker, P.~Glowacki, F.~Benmansour, and P.~Fua.
\newblock {Detecting Irregular Curvilinear Structures in Gray Scale and Color
  Imagery Using Multi-Directional Oriented Flux}.
\newblock In {\em International Conference on Computer Vision}, pages
  1553--1560, December 2013.

\bibitem{Turetken16a}
E.~Turetken, F.~Benmansour, B.~Andres, P.~Glowacki, H.~Pfister, and P.~Fua.
\newblock {Reconstructing Curvilinear Networks Using Path Classifiers and
  Integer Programming}.
\newblock {\em IEEE Transactions on Pattern Analysis and Machine Intelligence},
  38(12):2515--2530, 2016.

\bibitem{Wegner13}
J.D. Wegner, J.A. Montoya-Zegarra, and K.~Schindler.
\newblock {A Higher-Order CRF Model for Road Network Extraction}.
\newblock In {\em Conference on Computer Vision and Pattern Recognition}, pages
  1698--1705, 2013.

\bibitem{Wiedemann98}
C.~Wiedemann, C.~Heipke, H.~Mayer, and O.~Jamet.
\newblock {Empirical Evaluation of Automatically Extracted Road Axes}.
\newblock In {\em Empirical Evaluation Techniques in Computer Vision}, pages
  172--187, 1998.

\bibitem{Wolterink19}
J.~Wolterink, R.~van Hamersvelt, M.~Viergever, T.~Leiner, and I.~Isgum.
\newblock {Coronary Artery Centerline Extraction in Cardiac {CT} Angiography
  Using a Cnn-Based Orientation Classifier}.
\newblock {\em Medical Image Anal.}, 51:46--60, 2019.

\bibitem{Wu12a}
D.~Wu, D.~Liu, Z.~Puskas, C.~Lu, A.~Wimmer, C.~Tietjen, G.~Soza, and S.~K.
  Zhou.
\newblock {A Learning Based Deformable Template Matching Method for Automatic
  Rib Centerline Extraction and Labeling in CT Images}.
\newblock In {\em Conference on Computer Vision and Pattern Recognition}, 2012.

\end{thebibliography}
